\documentclass{article}
\usepackage{iclr2026_conference,times}

\usepackage{amsmath,amsfonts,bm}

\def\eqref#1{equation~\ref{#1}}

\def\1{\bm{1}}

\DeclareMathAlphabet{\mathsfit}{\encodingdefault}{\sfdefault}{m}{sl}
\SetMathAlphabet{\mathsfit}{bold}{\encodingdefault}{\sfdefault}{bx}{n}

\usepackage{hyperref}
\usepackage{url}
\usepackage{graphicx}
\usepackage{amsmath}   
\usepackage{amssymb}   

\title{Agentic Exploration of PDE Spaces using Latent Foundation Models for Parameterized Simulations}

\author{%
  Abhijeet Vishwasrao$^{*}$\thanks{$^{*}$Equal contribution.} \\
  University of Michigan
\And
  Francisco Giral$^{**}$ \\
  Universidad Politécnica de Madrid
\AND
  Mahmoud Golestanian \\
  Purdue University
\And
  Federica Tonti \\
  University of Michigan
\And
  Andrea Arroyo Ramo \\
  Universitat Politècnica de València
\AND
  Adrian Lozano-Duran \\
  Caltech
\And
  Steven L.~Brunton \\
  University of Washington
\And
  Sergio Hoyas \\
  Universitat Politècnica de València
\AND
  Soledad Le Clainche \\
  Universidad Politécnica de Madrid
\And
  Hector Gomez \\
  Purdue University
\And
  Ricardo Vinuesa \\
  University of Michigan
}

\iclrfinalcopy 
\begin{document}

\maketitle

\begin{abstract}
Flow physics and more broadly physical phenomena governed by partial differential equations (PDEs), are inherently continuous, high-dimensional and often chaotic in nature. Traditionally, researchers have explored these rich spatiotemporal PDE solution spaces using laboratory experiments and/or computationally expensive numerical simulations. This severely limits automated and large-scale exploration, unlike domains such as drug discovery or materials science, where discrete, tokenizable representations naturally interface with large language models. We address this by coupling multi-agent LLMs with \emph{latent foundation models} (LFMs), a generative model \emph{over parametrised simulations}, that learns explicit, compact and disentangled latent representations of flow fields, enabling continuous exploration across governing PDE parameters and boundary conditions. The LFM serves as an on-demand surrogate simulator, allowing agents to query arbitrary parameter configurations at negligible cost. A hierarchical agent architecture orchestrates exploration through a closed loop of hypothesis, experimentation, analysis and verification, with a tool-modular interface requiring no user support. Applied to flow past tandem cylinders at $Re = 500$, the framework autonomously evaluates over 1{,}600 parameter-location pairs and discovers divergent scaling laws: a regime-dependent two-mode structure for minimum displacement thickness and a robust linear scaling for maximum momentum thickness, with both landscapes exhibiting a dual-extrema structure that emerges at the near-wake to co-shedding regime transition. The coupling of the learned physical representations with agentic reasoning establishes a general paradigm for automated scientific discovery in PDE-governed systems.
\end{abstract}

\section{Introduction}

In fluid dynamics, scientific discovery has traditionally relied on systematic parameter studies: researchers design simulation campaigns, execute costly computations and manually interpret results to identify regimes, transitions and quantities of interest. While advances in computational fluid dynamics now enable high-fidelity simulations of complex flows, exhaustive exploration of parameter spaces remains prohibitively expensive. Data-driven reduced-order models (ROMs) and neural surrogates have emerged as efficient alternatives for rapid flow-field prediction \citep{vinuesa_enhancing_2022}, yet these tools typically serve as passive components within human-driven workflows, the task of deciding \emph{what} to simulate, \emph{where} to refine resolution and \emph{how} to synthesize findings still falls to the researcher.

Recent progress in large language models (LLMs) has demonstrated remarkable capabilities in multi-step reasoning \citep{wei_chain--thought_2022}, tool use \citep{schick_toolformer_2023} and agentic problem-solving \citep{yao_react_2022,shinn_reflexion_2023}. When augmented with external evaluators, such systems can achieve genuine discovery: FunSearch \citep{romera-paredes_mathematical_2024} and AlphaEvolve \citep{novikov_alphaevolve_2025} couple LLM-generated proposals with programmatic verification to surpass known results on combinatorial problems. However, these frameworks operate over discrete, symbolic domains, such as code, algorithms, mathematical expressions, where correctness can be automatically verified. Extending agentic discovery to physical systems governed by continuous spatiotemporal dynamics, where ``ground truth'' emerges from the structure of flow-fields  rather than symbolic evaluation, remains largely unexplored \citep{vinuesa_explainable_2026}.

We propose a framework that couples a multi-agent LLM system with a \emph{latent foundation model} (LFM), a generative model over parametrized simulation spaces, combining a total-correlation variational autoencoder with diffusion-based sampling, to enable systematic, autonomous exploration of parameterized flow physics. The LFM encodes a continuous manifold of physically plausible flow-fields , providing on-demand access to any point in parameter space without requiring new simulations. A hierarchical agent architecture orchestrates exploration: a \emph{Planner} formulates investigation strategies based on current knowledge, an \emph{Analyst} queries the LFM and computes interpretable flow statistics using physics-aware tools and a \emph{Critic} validates outputs before updating the knowledge base. This closed loop of hypothesis, experimentation, analysis and refinement enables the system to autonomously map physical regimes without hard-coded domain knowledge.

We demonstrate this framework on incompressible flow past tandem cylinders, a canonical configuration exhibiting rich wake-interaction phenomena \citep{sumner_two_2010,zhou_wake_2024, carmo_numerical_2006, sharman_numerical_2005}. Given only the task of characterizing the inter-cylinder wake structure, the agent autonomously samples cylinder spacings and computes integral thickness profiles across the parameter space. The agent discovers qualitatively divergent scaling laws: minimum displacement thickness follows a regime-dependent two-mode structure tied to the near-wake to co-shedding transition, while maximum momentum thickness follows a single linear law tracking the downstream cylinder, results that would require exhaustive manual parameter sweeps to uncover. The framework can be equipped with different analysis tools to extract additional quantities such as vortex shedding frequencies or aerodynamic coefficients. This work demonstrates that LLM-driven agents, when based on learned physical representations, can systematically explore and characterize complex flow physics with no human intervention.

\section{Methods}

\subsection{Latent Foundation Model over Parameterized Simulation Spaces}

We consider PDE systems based on two-dimensional, incompressible Navier--Stokes equations, where the governing equations are parameterized by a set of input parameters $\mathcal{P} \in \Theta \subset \mathbb{R}^{n_\mathcal{P}}$. The parameter space $\Theta$ may encode geometric configurations, boundary conditions or flow parameters. For each parameter configuration $\mathcal{P}_i$, numerical simulation yields spatiotemporal flow-fields  $\mathbf{x}^{(\mathcal{P})}(t) \in \mathbb{R}^{C \times H \times W}$, where $C$ denotes the number of physical variables and $H \times W$ is the spatial resolution. We assume access to a dataset $\mathcal{D} = \{(\mathbf{x}_i, \mathcal{P}_i)\}_{i=1}^{N}$ comprising snapshots from simulations across a discrete set of parameter configurations $\{\mathcal{P}_1, \ldots, \mathcal{P}_M\} \subset \Theta$. Our goal is to construct a lightweight latent foundation model over parameter space $\mathcal{P} \in \Theta$ that enables continuous sampling of statistically accurate and physically plausible flow-fields  enabling interpolation and moderate extrapolation across parameter space $\mathcal{P}$.


The latent foundation model (LFM) consists of two components: a total-correlation variational autoencoder (TC-VAE) that learns a compact, disentangled latent representation of the flow-fields  and a parameter-conditioned diffusion model that enables generative sampling across the parameter space $\mathcal{P}$.

\subsubsection{Total Correlation Variational Autoencoder}

Variational autoencoders (VAEs) learn latent representations by maximizing a variational lower bound on the data log-likelihood \citep{higgins_beta-vae_2017}. Given flow-field observations $\mathbf{x}$, we introduce latent variables $\mathbf{z} \in \mathbb{R}^{d_z}$ and optimize:
\begin{equation}
\log p(\mathbf{x}) \geq \mathcal{L}_{\text{ELBO}}(\mathbf{x}; \phi, \psi) = \mathbb{E}_{q_\phi(\mathbf{z}|\mathbf{x})}\left[\log p_\psi(\mathbf{x}|\mathbf{z})\right] - D_{\text{KL}}\left(q_\phi(\mathbf{z}|\mathbf{x}) \| p(\mathbf{z})\right),
\label{eq:elbo}
\end{equation}
where $q_\phi(\mathbf{z}|\mathbf{x}) = \mathcal{N}(\mathbf{z}; \boldsymbol{\mu}_\phi(\mathbf{x}), \mathrm{diag}(\boldsymbol{\sigma}^2_\phi(\mathbf{x})))$ is the encoder, $p_\psi(\mathbf{x}|\mathbf{z})$ is the decoder and $p(\mathbf{z}) = \mathcal{N}(\mathbf{0}, \mathbf{I})$ is the prior. The first term encourages reconstruction fidelity; the second regularizes the latent distribution. The $\beta$-VAE \citep{higgins_beta-vae_2017} introduces a hyperparameter $\beta > 1$ weighting the KL term to encourage disentanglement:
\begin{equation}
\mathcal{L}_{\beta\text{-VAE}} = \mathbb{E}_{q_\phi(\mathbf{z}|\mathbf{x})}\left[\log p_\psi(\mathbf{x}|\mathbf{z})\right] - \beta   D_{\text{KL}}\left(q_\phi(\mathbf{z}|\mathbf{x}) \| p(\mathbf{z})\right).
\label{eq:beta_vae}
\end{equation}
However, uniformly penalizing the KL term often introduces a trade-off between reconstruction quality and disentanglement. The total correlation VAE ($\beta$-TC-VAE) \citep{chen_isolating_2018} addresses this by decomposing the KL divergence into three interpretable components:
\begin{equation}
D_{\text{KL}}\left(q(\mathbf{z}|\mathbf{x}) \| p(\mathbf{z})\right) = \underbrace{I_q(\mathbf{x}; \mathbf{z})}_{\text{mutual information}} + \underbrace{D_{\text{KL}}\left(q(\mathbf{z}) \| \prod_{j} q(z_j)\right)}_{\text{total correlation}} + \underbrace{\sum_{j} D_{\text{KL}}\left(q(z_j) \| p(z_j)\right)}_{\text{dimension-wise KL}},
\label{eq:kl_decomposition}
\end{equation}
where $I_q(\mathbf{x}; \mathbf{z})$ quantifies information preserved in the latent code, the total correlation $\mathrm{TC}(\mathbf{z})$ measures statistical dependence among latent dimensions and the dimension-wise KL ensures that each marginal $q(z_j)$ remains close to the prior. The $\beta$-TC-VAE loss applies independent weights:
\begin{equation}
\mathcal{L}_{\text{TC-VAE}} = -\mathbb{E}_{q_\phi}\left[\log p_\psi(\mathbf{x}|\mathbf{z})\right] + \alpha   I_q(\mathbf{x}; \mathbf{z}) + \beta   \mathrm{TC}(\mathbf{z}) + \gamma   \sum_{j} D_{\text{KL}}\left(q(z_j) \| p(z_j)\right).
\label{eq:tc_loss}
\end{equation}
Setting $\alpha = 0$ preserves reconstruction fidelity, while $\beta \gg 1$ penalizes inter-dimension dependence, driving each latent dimension to capture independent factors of variation from the parameter space $\mathcal{P}$. 


\subsubsection{Parameter-Conditioned Latent Diffusion}

To enable generative sampling conditioned on the input parameters $\mathcal{P}$, we train a diffusion model in the learned latent space. Diffusion models define a forward process that gradually corrupts data with Gaussian noise and learn to reverse this process \citep{song_score-based_2020}. Given a latent representation $\mathbf{z}_0 \sim q(\mathbf{z})$, the forward process produces noisy samples:
\begin{equation}
\mathbf{z}_\sigma = \mathbf{z}_0 + \sigma \boldsymbol{\epsilon}, \quad \boldsymbol{\epsilon} \sim \mathcal{N}(\mathbf{0}, \mathbf{I}),
\label{eq:forward_diffusion}
\end{equation}
where $\sigma \in [\sigma_{\min}, \sigma_{\max}]$ indexes the noise level. Following the EDM framework \citep{karras_elucidating_2022}, we train a denoiser $D_\xi(\mathbf{z}_\sigma, \sigma, \mathcal{P})$ to recover the clean sample:
\begin{equation}
\mathcal{L}_{\text{DM}} = \mathbb{E}_{\mathbf{z}_0, \sigma, \boldsymbol{\epsilon}}\left[\lambda(\sigma) \left\| D_\xi(\mathbf{z}_0 + \sigma\boldsymbol{\epsilon}, \sigma, \mathcal{P}) - \mathbf{z}_0 \right\|_2^2\right],
\label{eq:diffusion_loss}
\end{equation}
where $\lambda(\sigma)$ is a noise-level-dependent weighting and $\mathcal{P}$ is the conditioning parameter. Network inputs and outputs are preconditioned following EDM conventions to ensure stable training across noise levels. The denoiser is parameterized as a transformer employing factorized spatiotemporal attention, decomposing into spatial and temporal components with rotary positional embeddings (RoPE) for temporal encoding, with parameter conditioning injected via adaptive layer normalization (adaLN) i.e. $\mathcal{P}$ is encoded through a multilayer perceptron and combined with the noise embedding to modulate scale and shift parameters within each transformer block.

For temporal coherence, we adopt an autoregressive formulation. Let $\mathbf{z}_{t-k:t-1}$ denote a context window of $k$ previous latent states. The model predicts the next state conditioned on this history:
\begin{equation}
p_\xi(\mathbf{z}_t | \mathbf{z}_{t-k:t-1}, \mathcal{P}) = \int p_\xi(\mathbf{z}_t | \mathbf{z}_\sigma, \mathbf{z}_{t-k:t-1}, \mathcal{P}) \, p(\mathbf{z}_\sigma) \, \mathrm{d}\mathbf{z}_\sigma,
\label{eq:autoregressive}
\end{equation}
enabling generation of arbitrarily long, temporally consistent trajectories. Sampling proceeds by iteratively denoising from $\mathbf{z}_{\sigma_{\max}} \sim \mathcal{N}(\mathbf{0}, \sigma_{\max}^2 \mathbf{I})$ to $\mathbf{z}_0$ using the learned denoiser, conditioned on the requested parameters $\mathcal{P}$.

The complete generative pipeline operates as: $\mathcal{P} \xrightarrow{\text{condition}} D_\xi \xrightarrow{\text{sample}} \mathbf{z}_{0:T} \xrightarrow{p_\psi} \mathbf{x}_{0:T}$, where the diffusion model generates latent trajectories and the VAE decoder reconstructs full-resolution flow-fields . This enables on-demand generation of flow-fields  for arbitrary $\mathcal{P} \in \Theta$, including interpolation between and extrapolation beyond training configurations.

\subsection{Multi-Agent Exploration Framework}

\begin{figure}[t]
\begin{center}
\includegraphics[width=\textwidth]{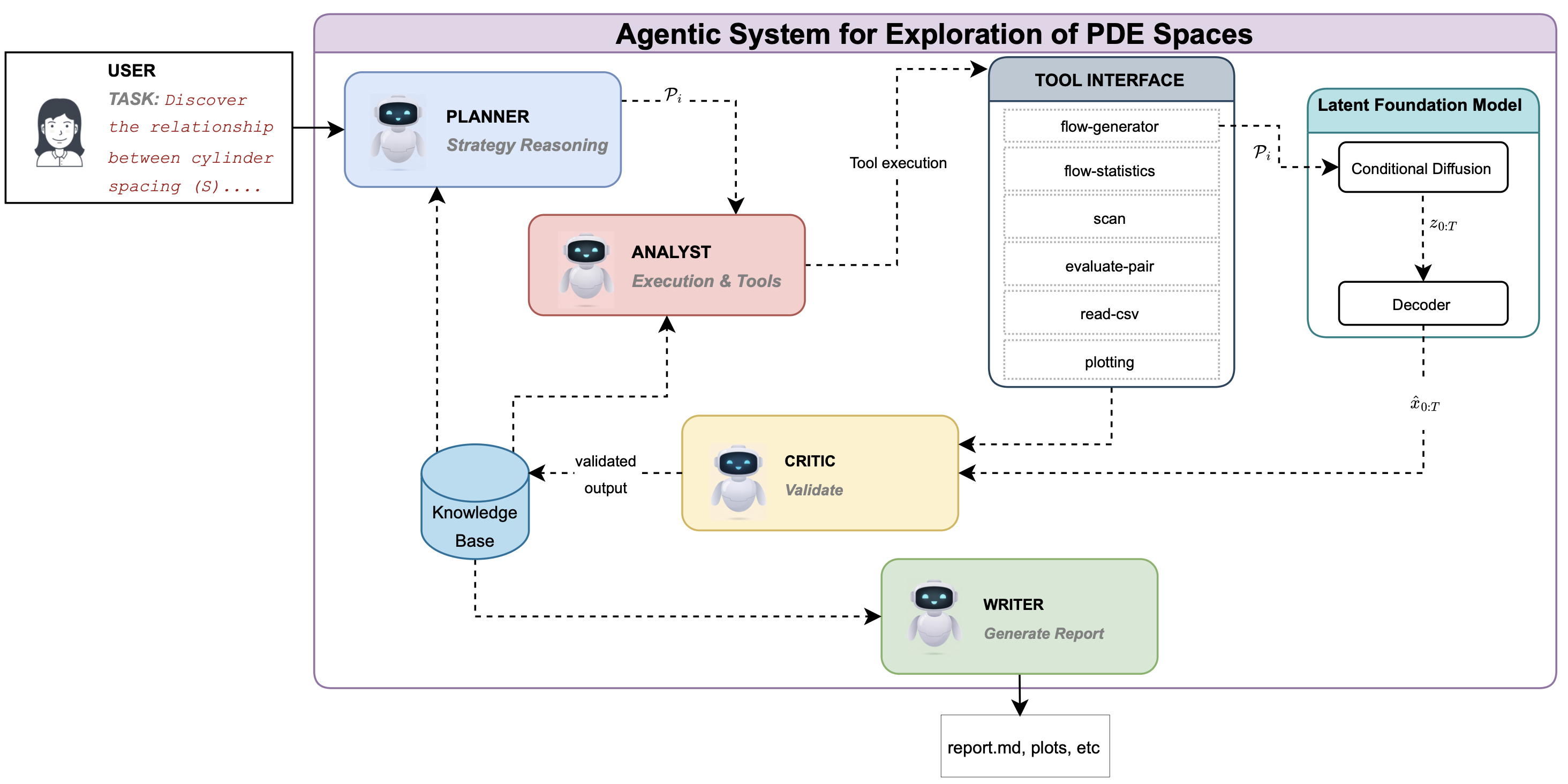}
\end{center}
\caption{Overview of the multi-agent exploration framework. A user-specified task initiates the exploration loop: the Planner formulates sampling strategies over the parameter space $\mathcal{P}$, the Analyst executes these through a modular tool interface that routes physics queries to the latent foundation model (LFM) for on-demand flow-field generation and the Critic validates outputs before updating the shared knowledge base. This closed loop of strategy, experimentation and verification continues until convergence criteria are met, after which the Writer synthesizes findings into a structured report.}
\label{fig:framework}
\end{figure}

To transform the LFM from a passive generative model into an active exploration system, we embed it within a multi-agent framework built on LangGraph. The framework implements a state machine where specialized agents, each an LLM instantiation equipped with role-specific prompts and tool access, collaborate to systematically explore the parameter space $\Theta$ in service of a user-specified task. Figure~\ref{fig:framework} illustrates the overall architecture.

\subsubsection{Task Specification and Agent Roles}

The exploration process begins with a user-provided task that defines the objective. This task specification guides the agents' reasoning without prescribing the exploration strategy. The framework comprises four agents with distinct responsibilities:

\paragraph{Planner.} The Planner serves as the strategic coordinator, maintaining a high-level view of exploration progress and formulating investigation strategies. Given the current knowledge state, a growing dataset of parameter configurations and associated flow statistics, the Planner reasons about which regions of parameter space require further investigation. In the present work it operates a coverage-based strategy, ensuring sufficient sampling across the full parameter range. The architecture naturally supports richer exploration strategies, for instance, targeted refinement near detected transitions, through prompt modification alone, without changes to the agent graph.

\paragraph{Analyst.} The Analyst executes the directives of the Planner as a deterministic execution node, interfacing directly with the LFM and physics-aware analysis tools. For each parameter configuration requested by the Planner, it: (i) queries the LFM to generate flow-field trajectories $\mathbf{x}_{0:T}$; (ii) computes interpretable flow statistics from the generated fields; and (iii) compiles results into structured formats for subsequent reasoning. By design, the Analyst requires no LLM reasoning, it translates the Planner's strategic decisions into systematic physical evaluations.

\paragraph{Critic.} The Critic provides quality control by validating the Analyst's outputs before they inform subsequent planning. As a first step, it verifies the geometric fidelity of each generated flow field by checking that the LFM-predicted cylinder center locations match the expected positions to within a tolerance; runs that fail this check are flagged as unreliable and excluded from the coverage assessment. Operating via a ReAct loop \citep{yao_react_2022}, it then reads the accumulated dataset via tool calls, evaluates coverage across the parameter space and assesses the smoothness of the emerging $x_p^*(\mathcal{S})$ trend. The Critic marks exploration as complete when all spacing windows contain at least one geometrically valid run or when the iteration budget is exhausted; otherwise it returns control to the Planner.

\paragraph{Writer.} Upon task completion, the Writer synthesizes the accumulated findings into a structured report. It reads the final dataset, generates summary visualizations and produces documentation that captures the exploration outcomes, including identified regimes, transition boundaries and optimal configurations.

\subsubsection{Tool Interface}

The agents access the LFM and analysis capabilities through a structured tool interface. The core tools include:

\begin{itemize}
    \item \texttt{flow\_generator($\mathcal{P}$)}: Queries the LFM to generate a temporally coherent flow-field trajectory for the specified parameter configuration.
    \item \texttt{scan($\mathcal{P}$)}: Performs a systematic sweep over analysis locations for a given configuration, computing error metrics at each point.
    \item \texttt{evaluate\_pair($\mathcal{P}$, $x_p$)}: Computes wake statistics including velocity-deficit profiles, momentum thickness and recovery metrics at a specified probe location.
    \item \texttt{read\_csv\_file(path)}: Retrieves accumulated results for reasoning and decision-making.
    \item \texttt{plot\_*}: Visualization tools for generating diagnostic figures.
\end{itemize}

These tools encapsulate domain expertise as callable functions, enabling the LLM agents to exploit physics-based analysis without requiring explicit physical knowledge in their prompts.

\subsubsection{Exploration Loop}

The agents operate in a closed-loop cycle that mirrors the scientific method (Figure~\ref{fig:framework}):

\begin{enumerate}
    \item \textbf{Task:} The user provides an exploration objective that defines success criteria and quantities of interest.
    \item \textbf{Hypothesis:} The Planner examines the current dataset and formulates a sampling strategy, initially a coarse sweep, later targeted refinement based on coverage gaps.
    \item \textbf{Experimentation:} The Analyst executes the strategy by querying the LFM for each requested $\mathcal{P}$, generating flow-fields  that would require expensive numerical simulation in traditional workflows.
    \item \textbf{Analysis:} The Analyst applies physics-aware tools to extract quantitative descriptors, building a structured dataset of parameter-observable pairs.
    \item \textbf{Verification:} The Critic reviews new data points against physical constraints and consistency checks. Valid results are accepted; problematic entries trigger re-analysis.
    \item \textbf{Refinement:} The Planner receives the validated dataset, updates its understanding of the parameter space and formulates the next hypothesis, identifying which coverage windows remain unfilled.
\end{enumerate}

This cycle continues until termination criteria are met, namely, error thresholds satisfied, sufficient coverage achieved or iteration budget exhausted. The Writer then generates a final report summarizing the exploration outcomes.

Crucially, the framework requires no predefined knowledge of what physical regimes exist or where transitions occur; these emerge from iterative interrogation of the parameter space. The LFM provides physical data by generating realistic flow-fields , while the agents provide reasoning capacity to interpret results, strategize sampling and systematically build understanding of the parameter-space structure.

\section{Results}
\label{sec:results}

We instantiate the framework on two-dimensional flow past tandem circular cylinders at $Re = 500$, based on the cylinder diameter $D$ and freestream velocity $U_\infty$. The parameterization reduces to a single scalar $\mathcal{P} \equiv \mathcal{S} \in [3.5, 10]$, the center-to-center cylinder spacing non-dimensionalized by $D$. The agent is tasked with characterizing the inter-cylinder wake structure using two integral quantities: the displacement thickness $\delta^*$ and the momentum thickness $\theta$,
\begin{equation}
\delta^*(x) = \int \left(1 - \frac{\bar{u}}{U_\infty}\right) \mathrm{d}y, \quad \theta(x) = \int \frac{\bar{u}}{U_\infty}\left(1 - \frac{\bar{u}}{U_\infty}\right) \mathrm{d}y,
\label{eq:bl_thicknesses}
\end{equation}
where $\bar{u}(x,y)$ is the time-averaged streamwise velocity. The displacement thickness quantifies the mass deficit in the wake, while $\theta$ quantifies the momentum deficit and is directly related to drag through the von K\'arm\'an momentum integral. For each spacing $\mathcal{S}$, the agent independently seeks the inter-cylinder location $x_p$ that minimizes $\delta^*$ (best mass-flux recovery) and maximizes $\theta$ (peak momentum deficit), yielding extremal locations $x_{\delta^*}^*(\mathcal{S})$ and $x_\theta^*(\mathcal{S})$, evaluated relative to local reference velocity ($U_{\infty}$). 

\subsection{Autonomous Exploration Behavior}

The framework executes two independent exploration campaigns, one per metric. The Planner operates a coverage-based strategy: the spacing domain $\mathcal{S} \in [3.5, 10.0]D$ is partitioned into uniform windows of width $1D$ and each iteration assigns one new spacing value to each uncovered window. For each suggested $\mathcal{S}$, the Analyst performs a full sweep over probe locations $x_p \in [0.5D, \mathcal{S} - 0.75D]$ at step size $0.15D$. The Critic then verifies the geometric fidelity of the generated flow fields; spacings that fail the geometry check are retried within the same window in the next iteration. Exploration terminates once all windows contain at least one geometrically valid spacing or the iteration budget is exhausted. This coverage-driven strategy ensures uniform sampling across the full parameter range; targeted refinement near detected transitions (e.g., adding intermediate spacings where $x_p^*$ jumps sharply between adjacent windows) is directly achievable by modifying the Planner prompt, without any architectural change.

Across both campaigns, the agent evaluates 1{,}600 parameter-location pairs spanning 41 spacing configurations (20 per $\delta^*$ campaign, 21 per $\theta$ campaign). Such systematic evaluation across the full inter-cylinder gap for each spacing is impractical in traditional workflows, where each configuration requires independent simulation setup, mesh generation, solver execution and manual post-processing.

\subsection{Divergent Scaling Laws}

\begin{figure}[t]
\begin{center}
\includegraphics[width=\textwidth,
  trim={0 2 0 41},
  clip
]{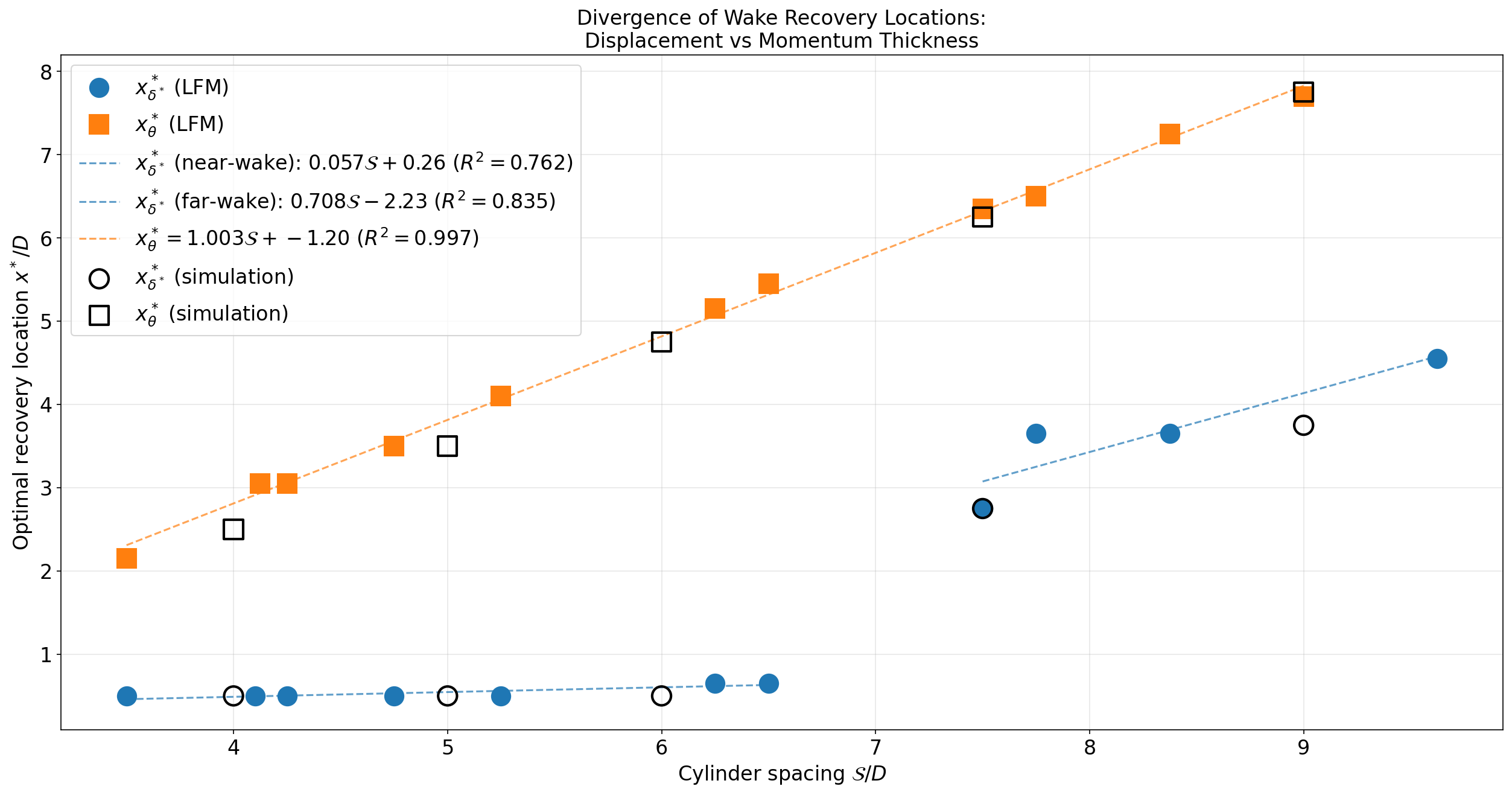}
\end{center}
\caption{Extremal locations $x_{\delta^*}^*$ (blue circles, minimum $\delta^*$) and $x_\theta^*$ (orange squares, maximum $\theta$) as functions of cylinder spacing $\mathcal{S}$. Filled markers: LFM-derived values; open markers: simulation ground truth. The minimum $\delta^*$ location transitions from a near-constant value ($\approx 0.5D$) in the near-wake regime to a linearly growing trend ($0.71\mathcal{S} - 2.2D$) in the co-shedding regime. The maximum $\theta$ location follows a single linear law ($\mathcal{S} - 1.2D$, $R^2 = 0.997$) across all spacings. The non-monotonic divergence between the two (shaded region) is a direct signature of the regime transition near $\mathcal{S} \approx 7D$.}
\label{fig:scaling}
\end{figure}

The agent discovers qualitatively distinct scaling behaviors for the two extremal locations (Figure~\ref{fig:scaling}). The maximum momentum thickness location $x_\theta^*$ follows a robust single linear law:
\begin{equation}
x_\theta^* \approx \mathcal{S} - 1.2D \quad (R^2 = 0.997),
\label{eq:scaling_theta}
\end{equation}
tracking the downstream cylinder position across all spacings. The minimum displacement thickness location $x_{\delta^*}^*$, however, reveals a \emph{regime-dependent} two-mode structure:
\begin{equation}
x_{\delta^*}^* \approx \begin{cases}
0.057\mathcal{S}+0.26D & \mathcal{S} \lesssim 6\text{--}7D \quad (R^2 = 0.76), \\
0.71\,\mathcal{S} - 2.2D & \mathcal{S} \gtrsim 7D \quad (R^2 = 0.84),
\end{cases}
\label{eq:scaling_delta}
\end{equation}
roughly corresponding to the experimentally validated interaction regimes of tandem cylinders \citep{sumner_two_2010, sharman_numerical_2005}.

\begin{figure}[t]
\begin{center}
\includegraphics[width=\textwidth]{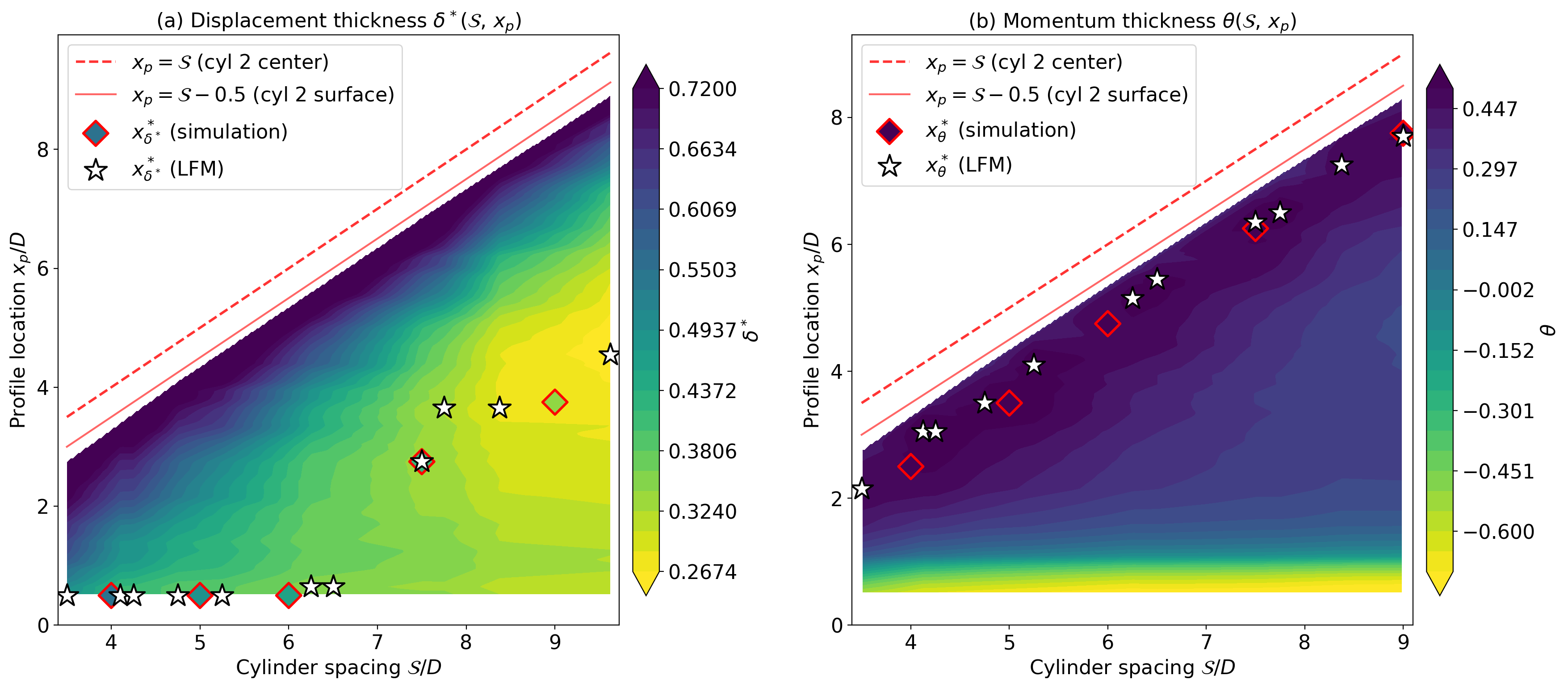}
\end{center}
\caption{Optimization landscape of \textbf{(a)} displacement thickness $\delta^*(\mathcal{S}, x_p)$ and \textbf{(b)} momentum thickness $\theta(\mathcal{S}, x_p)$ across the inter-cylinder gap. Stars: LFM-derived optima; red diamonds: simulation ground truth; dashed lines indicate the downstream cylinder center ($x_p = \mathcal{S}$) and surface ($x_p = \mathcal{S} - 0.5D$). \textbf{(a)} For small spacings a single minimum zone appears near $x_p \lesssim 1D$; for large spacings a second deeper minimum emerges at intermediate $x_p$, marking the near-wake to co-shedding transition near $\mathcal{S} \approx 7D$. \textbf{(b)} The global maximum tracks the downstream cylinder surface across all spacings, with a secondary local maximum visible near $x_p \approx 2D$ for higher $\mathcal{S}$.}
\label{fig:landscape}
\end{figure}

In the \emph{near-wake regime} ($\mathcal{S} \lesssim 6\text{--}7D$), the downstream cylinder lies within the upstream near-wake, suppressing independent vortex shedding. The minimum $\delta^*$ is anchored near $x_p \approx 0.5D$, immediately downstream of the upstream cylinder, because mass-flux recovery is governed by the near-wake structure of the upstream cylinder rather than the inter-cylinder distance. As $\mathcal{S}$ increases into the \emph{co-shedding regime} ($\mathcal{S} \gtrsim 7D$), both cylinders shed independently and a second, deeper minimum in $\delta^*$ emerges at larger $x_p$, displacing the global minimum downstream. This structural transition is directly visible in the displacement thickness landscape (Figure~\ref{fig:landscape}a): for small spacings, a single low-value zone appears near $x_p \lesssim 1D$; for large spacings, a second deeper minimum zone emerges at intermediate $x_p$ and the optimal locations (stars) transition from the near-upstream band to the mid-gap region.

The momentum thickness landscape (Figure~\ref{fig:landscape}b) reveals a complementary dual-extrema structure. A region of deeply negative $\theta$ at $x_p \lesssim 1D$ marks the recirculation zone behind the upstream cylinder, where reversed flow drives the momentum integrand strongly negative. Beyond this, $\theta$ rises to a first local maximum near $x_p \approx 1.5$--$2D$, then continues to increase for large spacings, reaching its global maximum at $x_p \approx \mathcal{S} - 1.2D$, immediately upstream of the downstream cylinder's stagnation zone. The optimal locations (stars) consistently track this near-cylinder maximum across all spacings, confirming the linear scaling of Eq.~(\ref{eq:scaling_theta}).

The divergence $|\Delta x^*| = |x_\theta^* - x_{\delta^*}^*|$ is consequently \emph{non-monotonic} (Table~\ref{tab:results}). In the near-wake regime, $x_{\delta^*}^*$ remains near $0.5D$ while $x_\theta^*$ grows linearly with $\mathcal{S}$, so the divergence increases steadily, reaching $\approx 5.3D$ at $\mathcal{S} = 7D$. At the regime transition, $x_{\delta^*}^*$ jumps from $\approx 0.5D$ to $\approx 2.75D$, sharply reducing the divergence to $\approx 3.5D$. Beyond the transition the divergence plateaus near $3.5\text{--}3.6D$, as the growth rates of $x_{\delta^*}^*$ and $x_\theta^*$ with spacing become comparable. This non-monotonic signature is a direct consequence of the regime transition: the abrupt jump in $x_{\delta^*}^*$ reflects the emergence of the second minimum in the $\delta^*$ landscape, with no corresponding discontinuity in $x_\theta^*$.

The two quantities are thus governed by fundamentally different physics: $x_{\delta^*}^*$ encodes the wake-interaction regime and transitions sharply with it, while $x_\theta^*$ tracks downstream cylinder proximity and is insensitive to regime. The divergence between them provides a quantitative signature of the regime state: growing in the near-wake regime, sharply reduced at transition and plateauing near $3.5D$ in the co-shedding regime. It must be noted that both Figure~\ref{fig:scaling} and Figure~\ref{fig:landscape} are generated by the agent utilizing available tools.

These findings carry direct engineering implications. Designs targeting drag reduction (governed by $\theta$ through the von K\'arm\'an momentum integral) should place sensing or control stations near the downstream cylinder face, independently of spacing. Designs targeting flow blockage reduction (governed by $\delta^*$) must account for the regime state: the optimal inter-cylinder location is fixed near the upstream cylinder in the near-wake regime but shifts downstream with spacing in the co-shedding regime, requiring fundamentally different design criteria in each case.

\begin{table}[t]
\caption{Extremal locations for minimum displacement thickness ($x_{\delta^*}^*$) and maximum momentum thickness ($x_\theta^*$) as discovered by the agent via LFM. A horizontal rule separates the near-wake and co-shedding regimes. The divergence $|\Delta x^*|$ grows steadily in the near-wake regime, drops sharply at the regime transition ($\mathcal{S} \approx 7D$), then plateaus near $3.5\text{--}3.6D$ in the co-shedding regime.}
\label{tab:results}
\begin{center}
\begin{tabular}{cccc}
\\ \hline \\
\multicolumn{1}{c}{\bf $\mathcal{S}/D$} & \multicolumn{1}{c}{\bf $x_{\delta^*}^*/D$} & \multicolumn{1}{c}{\bf $x_\theta^*/D$} & \multicolumn{1}{c}{\bf $|\Delta x^*|/D$} \\
\\ \hline \\
3.5 & 0.50 & 2.15 & 1.65 \\
4.5 & 0.50 & 3.05 & 2.55 \\
5.5 & 0.50 & 4.10 & 3.60 \\
6.5 & 0.65 & 5.45 & 4.80 \\
\hline
7.5 & 2.75 & 6.35 & 3.60 \\
8.5 & 3.65 & 7.25 & 3.60 \\
9.4 & 4.55 & 8.00 & 3.45 \\
\\ \hline \\
\end{tabular}
\end{center}
\end{table}

\subsection{Discussion}

In scientific domains such as mathematics \citep{romera-paredes_mathematical_2024}, drug discovery \citep{seal_ai_2025}, chemistry \citep{wadell_foundation_2025}, LLM-based agents operate on discrete, tokenizable representations, like SMILES strings, crystal graphs, amino acid sequences, etc. that naturally interface with language model architectures. Physical systems governed by partial differential equations present a fundamentally different challenge: they produce continuous, high-dimensional, chaotic spatiotemporal fields that cannot be meaningfully reduced to word tokens. Automating scientific exploration of PDE parameter spaces therefore requires an intermediate representation that makes the underlying physics accessible to LLM reasoning.

The latent foundation model provides this critical bridge. By learning \emph{explicit}, \emph{compact} and \emph{disentangled} latent representations of the high-dimensional flow-fields , the LFM enables continuous interpolation and controlled extrapolation across the governing parameters of the PDE. These three properties transform the LFM into an on-demand surrogate simulator: the agents can query arbitrary parameter configurations at negligible marginal cost relative to numerical simulation. This practically unlimited querying capacity is what enables the emergent adaptive behavior observed in our experiments. The tool-modular architecture of the framework further amplifies this capability. The analysis tools are decoupled from the agent architecture, so swapping the tool set changes what physics the agent can discover without modifying the agents themselves. No domain knowledge is assumed for the agent LLMs and physical insight emerges from the interplay of tool outputs and LLM reasoning, making the framework applicable to any parameterized PDE system for which an LFM can be trained.


For physical systems, exposing an LFM as a callable tool fundamentally alters the operating regime of the agent. Instead of treating each parameter configuration as an isolated, expensive experiment, the agents can densely probe the parameter space, cross-reference multiple physical diagnostics and adapt their exploration strategies in response to emerging structure. This capability has no analogue in traditional PDE workflows, where each query requires bespoke simulation setup and execution. In the tandem-cylinder case, 1,600 parameter–location evaluations across 41 spacing configurations reveal fine-scale structure in the optimization landscape that would be impractical to uncover through manual parameter sweeps.

\section{Conclusion}

We have presented a framework coupling multi-agent LLMs with a latent foundation model for autonomous exploration of parameterized PDE spaces. The key enabler is the explicit, compact and disentangled latent representation of the LFM, which makes continuous physical systems (high-dimensional, chaotic) accessible to LLM-driven scientific reasoning. Applied to tandem cylinder flow at $Re = 500$, the framework autonomously discovers divergent scaling laws, a regime-dependent two-mode structure for minimum displacement thickness and a robust linear scaling for maximum momentum thickness, alongside dual-extrema structures in both landscapes that emerge at the near-wake to co-shedding regime transition, findings that arise from exhaustive agentic evaluation rather than human-guided studies.

The framework opens several directions for future work. Extension to three-dimensional flows and multi-parameter spaces ($\dim(\Theta) > 1$) would test scalability to the high-dimensional parameter regimes common in scientific exploration and engineering design. Beyond guided exploration, the agents could autonomously formulate and evaluate their own hypotheses, proposing candidate physical mechanisms and designing targeted experiments to confirm or refute them, moving toward true agentic scientific discovery. Causal analysis within both latent and physical spaces would enable identification of mechanistic relationships rather than correlations alone. The tool-modular architecture readily accommodates spectral, force-decomposition and stability characterization as immediate extensions. More broadly, the coupling of learned physical representations with agentic reasoning establishes a general paradigm for automated scientific discovery in PDE-governed systems.

\bibliography{iclr2026_conference}
\bibliographystyle{iclr2026_conference}

\newpage
\appendix

\section{Latent Foundation Model: Architecture and Validation}
\label{app:lfm}

\subsection{Implementation Details}

This section provides concrete implementation details complementing the mathematical formulation in Section~2.1. The LFM comprises a $\beta$-TC-VAE encoder--decoder and a parameter-conditioned latent diffusion model; their formulations are given in Eqs.~\ref{eq:tc_loss}--\ref{eq:autoregressive}.

\paragraph{Dataset.} The model is trained on 89{,}614 snapshots of velocity $(u, v)$ and pressure $p$ fields at resolution $256 \times 128$, obtained from numerical simulations across 28 cylinder spacings $\mathcal{S}/D \in [3.5, 10]$. Data is partitioned by geometry: 14 spacings for training, 7 for validation and 7 for testing (approximately 50/25/25\%), ensuring the model is evaluated on entirely unseen geometric configurations rather than temporal interpolation within known spacings. Initial transients ($t < 250\,D/U_\infty$) are discarded.

\paragraph{TC-VAE.} The encoder uses four convolutional layers (stride 2, channels $3 \to 256$, SiLU activations) reducing spatial dimensions from $256 \times 128$ to $16 \times 8$, followed by projection to a two-dimensional latent space ($d_z = 2$). The decoder mirrors this structure with transposed convolutions. The TC-VAE loss weights (Eq.~\ref{eq:tc_loss}) are set to $\alpha = 0$, $\beta = 15$, $\gamma = 1$.

\paragraph{Latent diffusion.} The diffusion transformer processes sequences of latent states with $k = 2$ context frames for autoregressive generation. Geometry conditioning ($\mathcal{S}$) is injected via adaLN. Training follows the EDM framework \citep{karras_elucidating_2022}. Table~\ref{tab:architecture} summarizes key hyperparameters.

\begin{table}[h]
\caption{LFM architecture and training hyperparameters.}
\label{tab:architecture}
\begin{center}
\begin{tabular}{llll}
\\ \hline \\
\multicolumn{2}{c}{\bf TC-VAE} & \multicolumn{2}{c}{\bf Latent Diffusion} \\
\\ \hline \\
Latent dim.\ $d_z$ & 2 & Backbone & Transformer (DiT) \\
Encoder & 4 conv, stride 2 & Attention & Factorized (spatial + temporal) \\
Channels & 3 $\to$ 256 & Temporal encoding & RoPE \\
$(\alpha, \beta, \gamma)$ & $(0, 15, 1)$ & Conditioning & adaLN ($\mathcal{S}$ + $\sigma$) \\
Optimizer & Adam, lr $10^{-4}$ & Context frames $k$ & 2 \\
Batch size & 256 & Framework & EDM \\
Snapshots & 89{,}614 & Spacings (train/val/test) & 14 / 7 / 7 \\
\\ \hline \\
\end{tabular}
\end{center}
\end{table}

\subsection{Validation}

We validate the LFM at two levels: (i) field-level fidelity of the generated mean flow and Reynolds stress fields and (ii) accuracy of the derived integral quantities that the agent uses for discovery.

\begin{figure}[!h]
\begin{center}
\includegraphics[width=1\textwidth,
  trim={7 9 7 28},
  clip
]{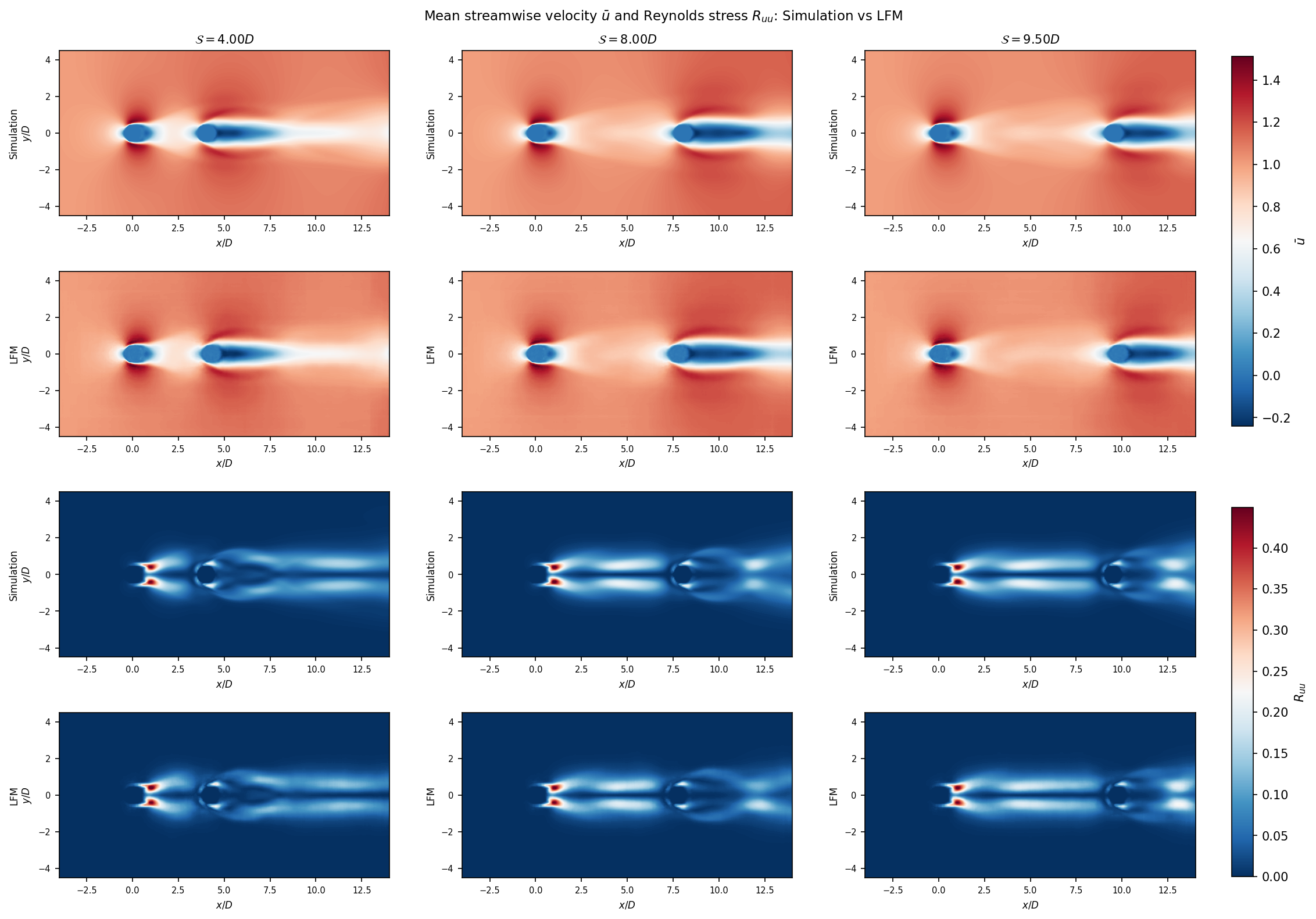}
\end{center}
\caption{Time-averaged streamwise velocity $\bar{u}$ (rows 1--2) and streamwise Reynolds stress $R_{uu}$ (rows 3--4): simulation (rows 1, 3) versus LFM (rows 2, 4) for $\mathcal{S}/D = 4, 8, 9.5$ (from test dataset). The LFM faithfully reproduces the mean wake topology, recirculation zones and wake-recovery structure, as well as the turbulent stress distribution, across the full spacing range.}
\label{fig:mean_validation}
\end{figure}

Figure~\ref{fig:mean_validation} compares the time-averaged streamwise velocity $\bar{u}$ and streamwise Reynolds stress $R_{uu}$ from direct numerical simulation and the LFM for three representative spacings spanning the parameter range ($\mathcal{S} = 4, 8, 9.5$). The LFM faithfully reproduces the mean wake topology, recirculation zones and wake-recovery structure across all configurations, including both the near-wake deficit behind the upstream cylinder and the interaction region approaching the downstream body. The Reynolds stress fields further confirm that the LFM captures the turbulent fluctuation structure in addition to the mean flow.

Beyond field-level reconstruction, we verify that the integral quantities central to the agentic exploration are accurately predicted. Figure~\ref{fig:landscape} (main text) overlays simulation ground truth (red diamonds) on the LFM-derived optimization landscapes. The LFM-derived optima closely track simulation for $\delta^*$ across all spacings and correctly reproduce the regime transition near $\mathcal{S} \approx 7D$. For $\theta$, the LFM captures the qualitative landscape structure and the proximity of the maximum to the downstream cylinder across all spacings, with minor quantitative deviations at some intermediate spacings. Overall, the surrogate preserves the physical landscape structure sufficiently for the agent to identify the divergent scaling laws and dual-extrema topology reported in Section~3.

\section{Multi-Agent Framework: Implementation Details}
\label{app:agents}

\subsection{LLM Backbone}

All LLM-based agents use Qwen2.5-32B-Instruct~\citep{qwen2025qwen25technicalreport}, a 32-billion-parameter open-weight model served locally via vLLM on a single A100 GPU. The model is accessed through an OpenAI-compatible API with temperature $0.35$ and a maximum token budget of $1{,}024$ per response. No commercial or closed-source models are used, the entire framework operates with locally hosted inference, demonstrating that the agentic exploration capabilities reported in this work do not require frontier-scale models.

\subsection{Agent Configurations}

Table~\ref{tab:agents} summarizes the reasoning mode, context inputs and output format for each agent. The Planner and Critic are the only agents that perform LLM reasoning, the Analyst is fully deterministic and the Writer uses a ReAct loop solely for tool orchestration.

\begin{table}[h]
\caption{Agent configurations. The Planner and Critic perform LLM reasoning, the Analyst executes deterministically.}
\label{tab:agents}
\begin{center}
\small
\begin{tabular}{llll}
\\ \hline \\
\textbf{Agent} & \textbf{Reasoning} & \textbf{Context inputs} & \textbf{Output} \\
\\ \hline \\
Planner & Direct LLM call & CSV summary, $x_p^*$ jump flags & JSON: \texttt{suggested\_ds}, \texttt{done} \\
Analyst & Deterministic & Planner's JSON & CSV rows (all $x_p$ per $\mathcal{S}$) \\
Critic & ReAct loop & CSV via tool call & JSON: \texttt{status}, \texttt{refinements} \\
Writer & ReAct loop & CSV + figures via tools & Markdown report \\
\\ \hline \\
\end{tabular}
\end{center}
\end{table}

\subsection{Prompt Structure}

We provide condensed versions of the Planner and Critic prompts, the two agents that perform scientific reasoning. 

\paragraph{Planner.} The Planner receives a system prompt specifying the metric under optimization (e.g., $\delta^*$ or $\theta$), the valid parameter range $\mathcal{S} \in [3.5, 10.0]D$ and a coverage strategy:
\begin{quote}
\small
\texttt{COVERAGE STRATEGY:} \\
\texttt{- Domain [3.5, 10.0]D divided into 1D windows} \\
\texttt{- Suggest one new S value per uncovered window} \\
\texttt{- Stop when all windows are covered or exhausted}
\end{quote}
At each iteration, the Planner receives the current window coverage status (covered/gap/exhausted) pre-computed from the CSV, together with the best $(x_p^*, \text{metric})$ per geometrically valid spacing. This context injection enables coverage-aware sampling without encoding domain-specific knowledge. Targeted refinement near detected transitions (e.g., adding intermediate spacings where $x_p^*$ jumps sharply between adjacent windows) can be activated by extending the Planner prompt with a refinement rule, without modifying the agent architecture.

\paragraph{Critic.} The Critic operates via a ReAct loop \citep{yao_react_2022} with access to a CSV reader tool. Its prompt specifies a geometric validity check followed by a coverage-based termination decision:
\begin{quote}
\small
\texttt{Decision Logic:} \\
\texttt{- Check cyl2\_error\_D for each new spacing} \\
\texttt{- Mark geometrically invalid spacings as unreliable} \\
\texttt{- If all windows COVERED or EXHAUSTED -> SOLVED} \\
\texttt{- If iteration >= max\_iterations -> SOLVED} \\
\texttt{- Otherwise -> NEEDS\_REFINEMENT (list gap windows)}
\end{quote}
The Critic reads the accumulated dataset via tool calls rather than receiving it in the prompt, ensuring decisions are grounded in actual data rather than the Planner's or Analyst's summaries.

\subsection{State Machine and Termination}

The framework is implemented as a LangGraph state machine with typed state comprising iteration count, accumulated messages, CSV path and exploration status. The node sequence is: Planner $\rightarrow$ Analyst $\rightarrow$ Critic, with the Critic performing both geometric validity assessment and coverage evaluation before routing. The routing logic after each Critic evaluation is:
\begin{itemize}
\item If \texttt{status = solved} and results CSV is non-empty $\rightarrow$ Writer
\item If iteration $\geq$ max iterations and results CSV is non-empty $\rightarrow$ Writer
\item Otherwise $\rightarrow$ Planner (next coverage cycle)
\end{itemize}
The two metric campaigns ($\delta^*$ and $\theta$) run as independent graph executions with separate state, CSV files and convergence criteria.

\subsection{Metric Modes and Objective Definition}
\label{app:metric_modes}

Although the experiments in Section~\ref{sec:results} focus on displacement thickness and momentum thickness separately, the implementation supports multiple objective definitions. Each mode defines: (i) the primary metric used for refinement and termination, (ii) a scalar convergence threshold and (iii) weights for a composite objective. The composite objective used internally by the tools is:
\begin{equation}
J = w_{\delta}\,\delta^*_{x_p} + w_{\theta}\,\theta_{x_p} + w_{L2}\,E_{L2} + w_{\cos}\,E_{\cos},
\label{eq:composite_objective}
\end{equation}
where $E_{L2}$ and $E_{\cos}$ are optional profile-matching errors against an upstream reference profile. The two campaigns reported in this paper correspond to:
\begin{itemize}
    \item \textbf{$\delta^*$ mode:} $(w_{\delta}, w_{\theta}, w_{L2}, w_{\cos}) = (1,0,0,0)$, primary metric $\delta^*_{x_p}$,
    \item \textbf{$\theta$ mode:} $(w_{\delta}, w_{\theta}, w_{L2}, w_{\cos}) = (0,1,0,0)$, primary metric $\theta_{x_p}$.
\end{itemize}
This modular design makes the framework extensible: new physical objectives can be introduced by adding new tool-computable metrics and registering a corresponding metric mode, without modifying the agent prompts or the state machine logic.

\subsection{Grounding and Anti-Fabrication Safeguards}
\label{app:grounding}

Because the Planner and Critic are LLM-based, the framework explicitly enforces grounding of all reported values in persisted artifacts. In particular, the Critic is implemented as a ReAct agent with access to \texttt{read\_csv\_file}. The Critic prompt requires that:

\begin{enumerate}
    \item all reported best values $(\mathcal{S}^*, x_p^*)$ are computed from the CSV via tool calls,
    \item termination decisions are made only after verifying convergence thresholds on the primary metric,
    \item refinement suggestions are justified by gaps in sampling density or detected discontinuities in $x_p^*(\mathcal{S})$.
\end{enumerate}

This design ensures that the Critic cannot ``accept'' fabricated values produced by Planner reasoning. If evidence is absent from the CSV, the Critic must return \texttt{needs\_refinement}, forcing further evaluation.

\subsection{Writer and Evidence Packaging}
\label{app:writer_impl}

The Writer agent is implemented as a tool-orchestrator rather than a free-form summarizer. Its prompt instructs it to call a single high-level report tool (\texttt{generate\_structured\_report}), which performs:

\begin{enumerate}
    \item ingestion of the final CSV and extraction of the best $x_p^*(\mathcal{S})$ per spacing,
    \item generation of all discovery figures (scaling laws, minimum-thickness curves and $(\mathcal{S},x_p)$ heatmaps),
    \item creation of a Markdown report containing the figures and CSV-derived tables.
\end{enumerate}

By generating tables and plots programmatically from the stored CSV, the Writer is prevented from introducing numerical values that are not supported by evidence. This final report step functions as an automated supplementary-material generator, producing a complete trace from the agent's exploration to the reported discovery.

\end{document}